\documentclass[10pt,twocolumn,letterpaper]{article}

\usepackage{cvpr} 
\usepackage{algorithm}
\usepackage{algorithmic}
\usepackage{multirow}
\usepackage{url}
\usepackage{bbding}
\usepackage{pifont}
\usepackage{wasysym}
\usepackage{amssymb}

\usepackage[dvipsnames]{xcolor}

\definecolor{cvprblue}{rgb}{0.21,0.49,0.74}
\usepackage[pagebackref,breaklinks,colorlinks,citecolor=cvprblue]{hyperref}

\def\paperID{3982} 
\def\confName{CVPR}
\def\confYear{2024}

\title{TranSegPGD: Improving Transferability of Adversarial Examples on Semantic Segmentation}

\author{Xiaojun Jia$^{1}$ , Jindong Gu$^{2}$, Yihao Huang$^{1}$,  Simeng Qin$^{3}$, Qing Guo$^{4}$, Yang Liu$^{1}$, Xiaochun Cao$^{5,\ddagger}$\\
$^{1}$Nanyang Technological University, Singapore  $^{2}$University of Oxford, UK\\        
$^{3}$ Yanshan University, China \\
$^{4}$ CFAR and IHPC, Agency for Science, Technology and Research (A*STAR), Singapore \\
$^{5}$ Shenzhen Campus of Sun Yat-sen University, China \\
{\tt\small jiaxiaojunqaq@gmail.com; jindong.gu@outlook.com; qinsm@stumail.ysu.edu.cn; } \\ {\tt\small huang.yihao@ntu.edu.sg; tsingqguo@ieee.org; yangliu@ntu.edu.sg; caoxiaochun@mail.sysu.edu.cn}
}

\begin{document}
\maketitle

\begin{abstract}

Transferability of adversarial examples on image classification has been systematically explored, which generates adversarial examples in black-box mode. However, the transferability of adversarial examples on semantic segmentation has been largely overlooked. In this paper, we propose an effective two-stage adversarial attack strategy to improve the transferability of adversarial examples on semantic segmentation, {dubbed} TranSegPGD.  Specifically, at the first stage, every pixel in an input image is divided into different branches based on its adversarial property. Different branches are assigned different weights for optimization to improve the adversarial performance of all pixels. 
We assign high weights to the loss of the hard-to-attack pixels to misclassify all pixels. At the second stage, the pixels are divided into different branches based on their transferable property which is dependent on Kullback-Leibler divergence. Different branches are assigned different weights for optimization to improve the transferability of the adversarial examples. We assign high weights to the loss of the high-transferability pixels to improve the transferability of adversarial examples. 
Extensive experiments with various segmentation models are conducted on PASCAL VOC
2012 and Cityscapes datasets to demonstrate the effectiveness of the proposed method. The proposed adversarial attack method can achieve state-of-the-art performance. 
 
\end{abstract}
\section{Introduction}
\begin{figure}[t]
\begin{center}
 \includegraphics[width=1.0\linewidth]{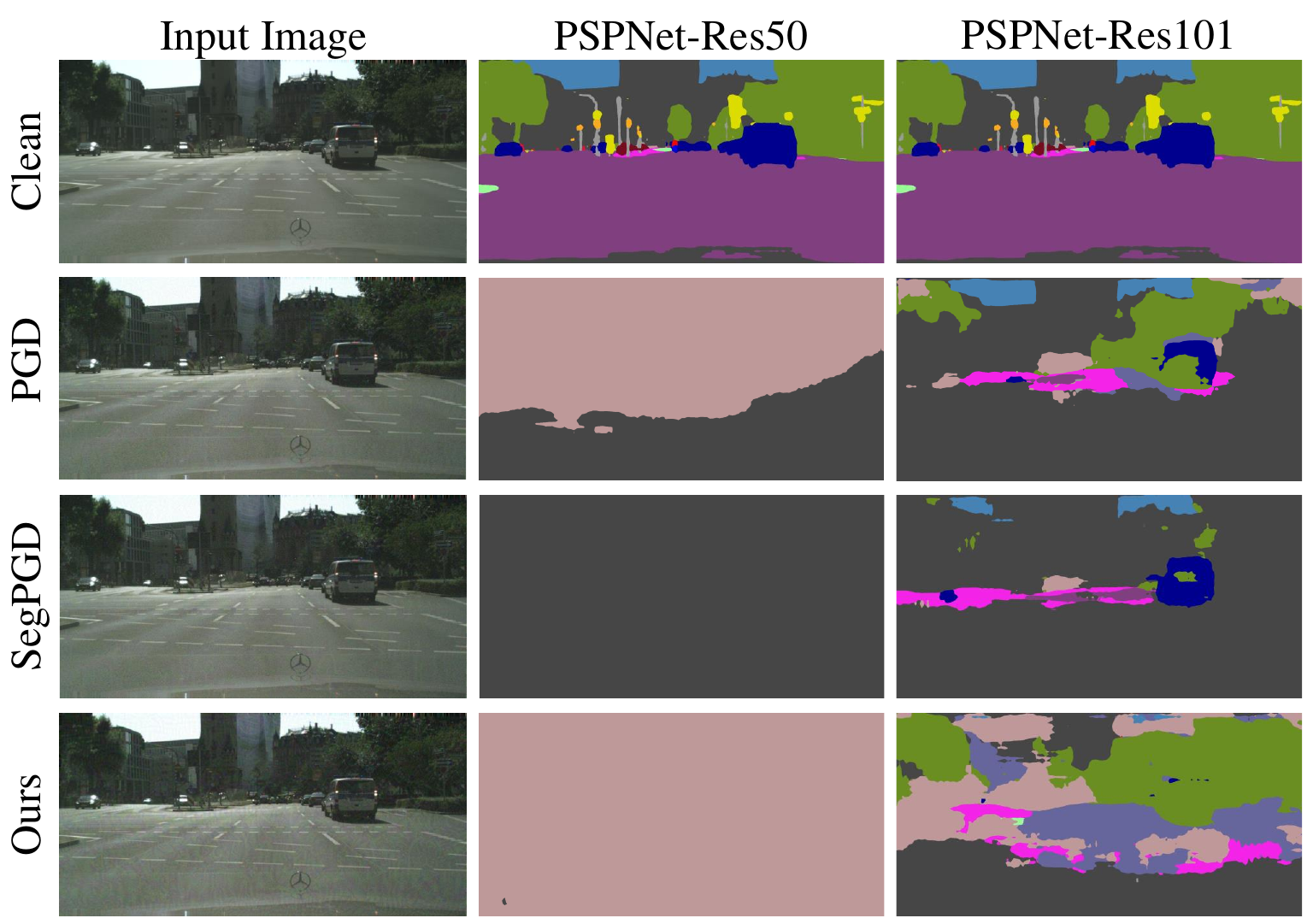}
\end{center}
\vspace{-3mm}
  \caption{ Visualization of Clean Images, Adversarial examples, and Segmentation Predictions. PSPNet-Res50 is used as the
source model and PSPNet-Res101 is used as the
target model.
The adversarial example generated by using the proposed method transfers better to other segmentation models. More figures are presented in the supplementary material. }
\label{fig:home}
\end{figure}
\begin{figure*}[t]
\begin{center}
 \includegraphics[width=1.0\linewidth]{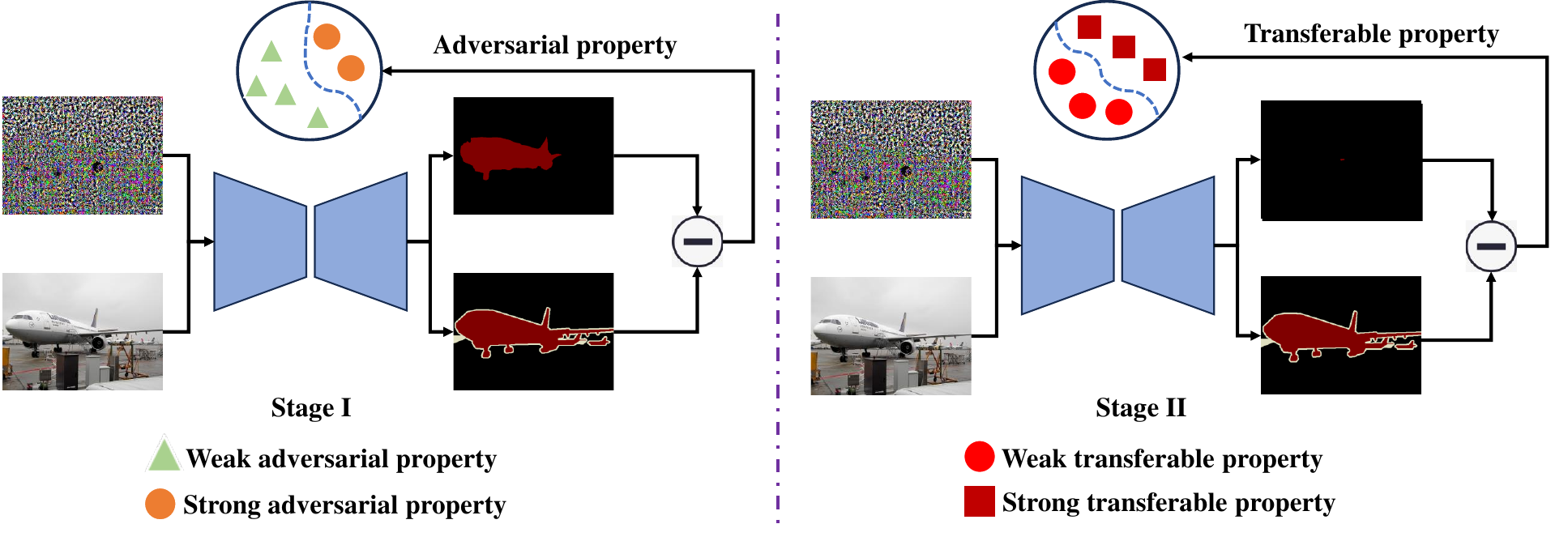}
\end{center}
\vspace{-3mm}
  \caption{ The proposed method consists of two-stage adversarial attack strategies. At the first stage, all pixels in the input image are divided into different branches based on their adversarial property. At the second stage, all pixels are divided into different branches based on their transferable property. }
\label{fig:framework}
\end{figure*}
A series of research works have indicated deep learning methods~\cite{lecun2015deep} are vulnerable to adversarial examples~\cite{DBLP:journals/corr/GoodfellowSS14, jia2020adv,he2023generating}, which are generated by adding the well-designed and imperceptible perturbations to the benign samples. Recent works adopt the generation of adversarial examples to study adversarial robustness in many fields of research~\cite{bai2019hilbert,jia2022adversarial,bai2021improving,bai2021clustering,wang2023adversarial}, such as speech recognition, image compression, and image generation, etc. In particular, many researchers focus on the task of image classification and generate adversarial examples to attack classification models through different perspectives~\cite{zhang2023improving,zhu2023boosting,huang2023erosion,bai2023query}. More importantly, several works have indicated that adversarial examples generated on a specific white-box classification model (source models) could also fool other different black-box classification models (target models), which can be considered as the transferability of adversarial examples~\cite{wu2020towards, bai2020improving, wu2021improving,wang2023role}. The concept of transferability in adversarial examples has garnered significant research attention due to its enabling of practical black-box attacks. In detail, they explore improving the transferability of adversarial examples on the image classification task from the perspectives of data augmentation~\cite{zhou2016learning,lin2019nesterov}, optimization methods~\cite{li2020towards,qin2022boosting}, and loss functions~\cite{zhao2021success,zhang2022investigating}. 

\par Although the transferability of adversarial examples generated on image classification tasks has been profoundly and comprehensively explored~\cite{gu2023survey,yu2023reliable}, the transferability of adversarial examples on semantic segmentation tasks, which can be regarded as an extension of the image classification task, has rarely been studied. In detail, image semantic segmentation endeavors to meticulously classify every individual pixel within the input image. Segmentation models have a wide range of applications in the real world, such as medical image segmentation. Hence, recent works~\cite{arnab2018robustness,xiao2018characterizing,frangi2018generalizability,he2019non,fischer2021scalable} pay much attention to the adversarial robustness of the image segmentation models. However, previous works~\cite{xu2021dynamic,gu2022segpgd} mainly focus on improving the success rate of adversarial examples on the white-box models, while ignoring the improvement of transferability. Some works~\cite{he2023transferable} even have found that it is hard for the adversarial examples generated on segmentation models to transfer across other segmentation models. As shown in Figure~\ref{fig:home}, previous works~\cite{xu2021dynamic,gu2022segpgd} about segmentation attack achieves limited transferable performance.

\par To improve the transferability of adversarial examples on semantic segmentation, we propose an effective two-stage adversarial attack strategy, \emph{dubbed} TranSegPGD. As shown in Figure~\ref{fig:framework}, we divide the entire generation process of adversarial examples on semantic segmentation into two stages. In detail, at the first stage, we divide every pixel of the input image into different branches based on its adversarial property. Then, we assign different weights to the different branches in the loss function for optimization to generate adversarial examples.
At the first stage, 
to misclassify all pixels of the input image, we assign high weights to the loss of the hand-to-attack pixels. Motivated by the related works of out-of-distribution~\cite{nagarajan2020understanding,wald2021calibration,zhu2022toward}, it is hard for well-trained models to classify the out-of-distribution examples. It indicates that adversarial examples that are farther distributedly from the original clean examples could have higher transferability. Generalized to semantic segmentation tasks, the generated adversarial pixels, which are farther distributedly from the original clean pixels, could have higher transferability. To improve the transferability of adversarial examples on semantic segmentation, they need to be assigned high weights during the second stage of adversarial example generation. Hence at the second stage, we first compute the Kullback-Leibler (KL) divergence, which can used to measure the distance between two distributions, for each pixel in the generated adversarial image with its corresponding pixel in the original clean image. Then we divided the pixels of the generated adversarial image into different branches based on its transferable property. The loss of the high-transferability pixels is assigned to high weights to improve adversarial transferability. As shown in Figure~\ref{fig:home}, the proposed method achieves better transferability than previous segmentation adversarial attack methods. 
\par Our main contributions are in three aspects:
\begin{itemize}
     \item We propose an effective two-stage adversarial attack strategy to improve the transferability of adversarial examples on semantic segmentation, \emph{dubbed} TranSegPGD.

    \item We also propose a dynamic attack step size to increase the diversity of generated adversarial examples, thus boosting the adversarial transferability. 

    \item Experiments and analyses across various network architectures and datasets are conducted to demonstrate the effectiveness of the proposed method. The proposed adversarial attack method can achieve state-of-the-art performance.
\end{itemize}

\section{Related Work}
\noindent \textbf{Adversarial attack on image classification:} for a given image classification network $f_{\boldsymbol{\theta}}(\cdot)$ with model parameters $\boldsymbol{\theta}$, the input data $\mathbf{x}$ and the corresponding ground truth label $\mathbf{y}$, the adversarial attack methods~\cite{DBLP:journals/corr/GoodfellowSS14,moosavi2016deepfool,madry2018towards,huang2023robustness} adopt the maximization of loss function $\mathcal{L}(f_{\boldsymbol{\theta}}(\mathbf{x}), \mathbf{y})$ to generate adversarial perturbations $\boldsymbol{\delta}$. In detail, Goodfellow \emph{et al.} propose to adopt Fast Gradient Sign Method
(FGSM)~\cite{DBLP:journals/corr/GoodfellowSS14}, which is an efficient gradient-based adversarial attack method, to generate adversarial examples. Madry \emph{et al.} use Projected Gradient Descent (PGD)~\cite{madry2018towards}, which is a multi-step iteration of FGSM, for adversarial example generation. Several works improve the attack performance of adversarial examples from multiple perspectives. Besides, recent works pay attention to improving the transferability of adversarial examples, which indicates the ability of adversarial examples generated on the white-box model to attack another black-box model. Specifically, a series of works adopt the data augmentation-based adversarial attack methods~\cite{wu2021improving,wang2021admix,huang2021transferable} to improve the adversarial transferability. For example, Xie \emph{et al.} ~\cite{xie2019improving} propose to perform I-FGSM with Diverse Inputs Method (DI-FGSM) to increase adversarial transferability. Dong \emph{et al.}~\cite{dong2019evading} propose to adopt Translation Invariance Method to implement I-FGSM (TI-FGSM) to enhance the transferability. Some works~\cite{wang2021enhancing,xiong2022stochastic} boost the transferability of the adversarial examples by using optimization-based methods. Dong \emph{et al.} ~\cite{dong2018boosting} use a Momentum Iterative Fast Gradient Sign Method (MI-FGSM), which is a classic adversarial attack method, for the improvement of transferability. Lin~\emph{et al.}~\cite{lin2019nesterov} propose Nesterov Iterative Fast Gradient Sign Method (NI-FGSM), which performs I-FGSM with Nesterov Accelerated Gradient (NAG), to boost transferability. 

\noindent \textbf{Adversarial attack on semantic segmentation:} previous works adopt adversarial attack methods to evaluate semantic segmentation models' robustness on the adversarial examples. In detail, Arnab \emph{et al.}~\cite{arnab2018robustness} propose to make use of FGSM and PGD to study the adversarial robustness of the semantic segmentation models. Some works \cite{hendrik2017universal,fischer2017adversarial} attack the semantic segmentation models by generating universal adversarial perturbations. Recently, some researchers~\cite{gu2022segpgd} have focused on exploring how to generate adversarial examples of semantic segmentation more efficiently. For example, Gu \emph{et al.}~\cite{gu2022segpgd} find that wrongly classified pixels, which drive the process of adversarial examples, cause an imbalanced gradient contribution, resulting in limited attack performance of adversarial examples. Then they propose an efficient adversarial attack method on semantic segmentation, which assigns high weights to the loss of the accurately classified pixels to relieve the impact of the wrongly classified pixels. They further propose to use the proposed attack method in adversarial training to improve model robustness, which mainly
focuses on how to improve the adversarial robustness of the model rather than improving the adversarial example transferability. 
Moreover, a series of works~\cite{xie2017adversarial,gu2021adversarial,he2023transferable} indicate that adversarial examples generated on semantic segmentation can easily overfit the source model, which makes the generated adversarial examples fool the target model hard. To improve the transferability of adversarial examples on semantic segmentation, we propose a two-stage adversarial attack strategy.

\section{The Proposed Method} 
We propose an effective two-stage adversarial attack adversarial attack strategy. In this section, we first introduce the pipeline of the proposed method. We introduce the first-stage adversarial example generation strategy. Then we propose a second-stage adversarial example generation strategy to improve transferability. 

\subsection{The pipeline of the proposed method}
The pipeline of the proposed method is shown in Fig.~\ref{fig:framework}. The proposed method divides the generation process of adversarial examples into two stages. During the first stage, the proposed method aims to improve the adversarial performance of adversarial examples on semantic segmentation. During the second stage, the proposed method aims to improve the adversarial transferability of adversarial examples on semantic
segmentation. 

\par For a given input benign image $\mathbf{x} \in \mathbb{R}^{H \times W \times C}$ and the corresponding segmentation label $\mathbf{y} \in \mathbb{R}^{H \times W \times M}$, a semantic segmentation model $f^{seg}_{\boldsymbol{\theta}} (\cdot)$ with the model parameters $\boldsymbol{\theta}$ categorizes every individual pixel $f^{seg}_{\boldsymbol{\theta}} (\mathbf{x}) \in \mathbb{R}^{H \times W \times M}$, where $H \times W$ represents the image size, $C$ represents the channel number of the input image, and $M$ represents the number of image classes. The objective of adversarial attacks on semantic segmentation is to generate an adversarial example, which can fool the segmentation model to misclassify the category of each pixel in the input image. However previous works mainly pay attention to adversarial performance on the source segmentation model, which is used to generate adversarial examples. However, they ignore the performance of the target model, which is not accessible to attackers. In this paper, we not only focus on the adversarial performance of the source model but also the adversarial performance of the target model, that is, how to improve the transferability of adversarial examples.
Specifically, we propose an effective two-stage adversarial attack strategy to improve adversarial transferability. The proposed method improves the adversarial performance of the generated adversarial examples at the first stage. The proposed method boosts the adversarial transferability of the generated adversarial examples at the second stage. 

\begin{algorithm*}[t]
\caption{Two-Stage Adversarial Attack Strategy}
\label{alg:attack}
\begin{algorithmic}[1] 
\REQUIRE
The semantic segmentation model $f^{seg}_{\boldsymbol{\theta}} (\cdot)$, the benign image $\mathbf{x}$, the corresponding label $\mathbf{y}$, the image size $H \times W$, the maximal perturbation $\epsilon$, the step size $\alpha$, and the iteration number $N$.
\STATE  $\mathbf{x}_{adv}^{0}=\mathbf{x}+\mathcal{U}(-\epsilon,+\epsilon)  \hfill \quad \triangleright$ initialization of the adversarial example
\FOR{$t=1,...,N$}    
\STATE  $P=f^{seg}_{\boldsymbol{\theta}} (\mathbf{x}_{adv}^{t})  \hfill \quad \triangleright$  Segmentation result on the adversarial example
\STATE $P_T, P_F \leftarrow P  \hfill \quad \triangleright$ The first stage of pixel division
\IF {$P_F \neq \varnothing $}                        
\STATE 
$\mathcal{L}\left(f^{seg}_{\boldsymbol{\theta}} \left(\mathbf{x}_{adv}^{t}\right), \mathbf{y}\right) = \frac{1-\alpha}{H \times W} \sum_{i \in P_T} \mathcal{L}_i\left(f^{seg}_{\boldsymbol{\theta}} \left(\mathbf{x}_{adv}^{t}\right), \mathbf{y}\right) +\frac{\alpha}{H \times W} \sum_{j \in P_F} \mathcal{L}_j\left(f^{seg}_{\boldsymbol{\theta}} \left(\mathbf{x}_{adv}^{t}\right), \mathbf{y}\right) \hfill \quad \triangleright $ Loss calculation
\ELSE
\STATE $D_{KL}(\mathbf{x}_{adv}) = \sum_{j=1}^n \sigma( f^{seg}_{\boldsymbol{\theta}}(\mathbf{x}_{adv}))_j \log \frac{\sigma( f^{seg}_{\boldsymbol{\theta}}(\mathbf{x}_{adv}))_j}{\sigma( f^{seg}_{\boldsymbol{\theta}}(\mathbf{x}))_j} \hfill \quad \triangleright$ KL distance of adversarial example
\STATE $P_L, P_H \leftarrow P  \hfill \quad \triangleright$ The second stage of pixel division
\STATE $\mathcal{L}\left(f^{seg}_{\boldsymbol{\theta}} \left(\mathbf{x}_{adv}^{t}\right), \mathbf{y}\right) = \frac{1-\beta}{H \times W} \sum_{i \in P_H} \mathcal{L}_i\left(f^{seg}_{\boldsymbol{\theta}} \left(\mathbf{x}_{adv}^{t}\right), \mathbf{y}\right) +\frac{\beta}{H \times W} \sum_{j \in P_L} \mathcal{L}_j\left(f^{seg}_{\boldsymbol{\theta}} \left(\mathbf{x}_{adv}^{t}\right), \mathbf{y}\right)  \hfill \quad \triangleright$ Loss calculation
\ENDIF
\STATE $\mathbf{x}_{adv}^{t+1}=\prod_{[-\epsilon, \epsilon]}\left[\mathbf{x}_{adv}^{t}+\alpha \cdot \operatorname{sign}\left(\nabla_{\mathbf{x}_{adv}^{t}} \mathcal{L}\left(f^{seg}_{\boldsymbol{\theta}} \left(\mathbf{x}_{adv}^{t}\right), \mathbf{y}\right)\right)\right]  \hfill \quad \triangleright$  Generation of adversarial examples
\ENDFOR
\end{algorithmic}
\end{algorithm*}
\subsection{The first-stage adversarial attack strategy}
During the first stage, the goal is to generate the adversarial example $\mathbf{x}_{adv}$ to misclassify each pixel of the input image as soon as possible. Previous works mainly adopt a classic adversarial attack method PGD to generate adversarial examples for semantic segmentation. It can be calculated as:
\begin{equation}
\mathbf{x}_{adv}^{t+1}=\prod_{[-\epsilon, \epsilon]}\left[\mathbf{x}_{adv}^{t}+\alpha \cdot \operatorname{sign}\left(\nabla_{\mathbf{x}_{adv}^{t}} \mathcal{L}\left(f^{seg}_{\boldsymbol{\theta}} \left(\mathbf{x}_{adv}^{t}\right), \mathbf{y}\right)\right)\right], 
\end{equation}
where $\mathbf{x}_{adv}^{t+1}$ represents the generated adversarial example after the $(t+1)$-th step, $\epsilon$ represents the maximum perturbation strength, $\alpha$ represents the attack step size, and $\mathcal{L}\left(f^{seg}_{\boldsymbol{\theta}} \left(\mathbf{x}_{adv}^{t}\right), \mathbf{y}\right)$ represents the cross-entropy loss. However, this approach assigns equal importance to every pixel, leading to a situation where misclassified pixels dominate the adversarial example generation process, thus limiting the adversarial performance of PGD on semantic segmentation. 
\par To generate adversarial examples more effectively during the first stage, following 
the previous work~\cite{gu2022segpgd}, we divide all pixels into two two branches based on the prediction accuracy, \emph{ i.e.,} the correctly classified pixels $P_{T}$ and the incorrectly classified pixels $P_{F}$. The loss function can be formulated:
\begin{equation}
\begin{aligned}
\mathcal{L}\left(f^{seg}_{\boldsymbol{\theta}} \left(\mathbf{x}_{adv}^{t}\right), \mathbf{y}\right) &= \frac{1-\gamma}{H \times W} \sum_{i \in P_T} \mathcal{L}_i\left(f^{seg}_{\boldsymbol{\theta}} \left(\mathbf{x}_{adv}^{t}\right), \mathbf{y}\right) \\ &+\frac{\gamma}{H \times W} \sum_{j \in P_F} \mathcal{L}_j\left(f^{seg}_{\boldsymbol{\theta}} \left(\mathbf{x}_{adv}^{t}\right), \mathbf{y}\right),
\end{aligned}
\end{equation}
where $\mathcal{L}_i$ represents the cross-entropy loss of the $i$-th pixel for semantic segmentation, and $\gamma$ represents a hyper-parameter. Although this method can effectively improve the adversarial performance of adversarial examples, it has limited improvement in adversarial transferability. Specifically, when all pixels are misclassified, the first-stage attack method still treats all pixels equally and ignores the different transferable properties of different pixels. 

\begin{table*}[t]
\centering
 \scalebox{0.85}{
\begin{tabular}{@{}ccc|c|c|c|c@{}}
\toprule
\multicolumn{3}{c|}{Target Models}                                                                                     & PSPNet-Res50   & PSPNet-Res101  & DeepLabV3-Res50 & DeepLabV3-Res101 \\ \midrule
\multicolumn{3}{c|}{Clean Images}                                                                                      & 78.55          & 79.11          & 78.17           & 80.55            \\ \midrule \midrule
\multicolumn{1}{c|}{\multirow{16}{*}{Source Models}} & \multicolumn{1}{c|}{\multirow{4}{*}{PSPNet-Res50}}     & PGD    & 4.60           & 36.91          & 5.05            & 38.59            \\ \cmidrule(l){3-7} 
\multicolumn{1}{c|}{}                                & \multicolumn{1}{c|}{}                                  & SegPGD & 2.09           & 36.76          & 2.42            & 38.31            \\ \cmidrule(l){3-7} 
\multicolumn{1}{c|}{}                                & \multicolumn{1}{c|}{}                                  & TranSegPGD (Ours)   & \textbf{1.55}  & \textbf{34.57} & \textbf{2.38}   & \textbf{36.74}   \\ \cmidrule(l){2-7} 
\multicolumn{1}{c|}{}                                & \multicolumn{1}{c|}{\multirow{4}{*}{PSPNet-Res101}}    & PGD    & 31.67          & 2.88           & 31.18           & 5.42             \\ \cmidrule(l){3-7} 
\multicolumn{1}{c|}{}                                & \multicolumn{1}{c|}{}                                  & SegPGD & 30.44          & 1.36           & 29.97           & 3.59             \\ \cmidrule(l){3-7} 
\multicolumn{1}{c|}{}                                & \multicolumn{1}{c|}{}                                  & TranSegPGD (Ours)   & \textbf{29.06} & \textbf{1.08}  & \textbf{27.88}  & \textbf{3.19}    \\ \cmidrule(l){2-7} 
\multicolumn{1}{c|}{}                                & \multicolumn{1}{c|}{\multirow{4}{*}{DeepLabV3-Res50}}  & PGD    & 4.04           & 32.49          & 3.72            & 33.81            \\ \cmidrule(l){3-7} 
\multicolumn{1}{c|}{}                                & \multicolumn{1}{c|}{}                                  & SegPGD & 2.38           & 31.43          & 1.63            & 33.25            \\ \cmidrule(l){3-7} 
\multicolumn{1}{c|}{}                                & \multicolumn{1}{c|}{}                                  & TranSegPGD (Ours)   & \textbf{2.10}  & \textbf{30.59} & \textbf{1.55}   & \textbf{30.97}   \\ \cmidrule(l){2-7} 
\multicolumn{1}{c|}{}                                & \multicolumn{1}{c|}{\multirow{4}{*}{DeepLabV3-Res101}} & PGD    & 31.23          & 4.84           & 30.67           & 3.48             \\ \cmidrule(l){3-7} 
\multicolumn{1}{c|}{}                                & \multicolumn{1}{c|}{}                                  & SegPGD & 30.62          & 2.89           & 30.15           & 1.58             \\ \cmidrule(l){3-7} 
\multicolumn{1}{c|}{}                                & \multicolumn{1}{c|}{}                                  & TranSegPGD (Ours)   & \textbf{29.93} & \textbf{2.73}  & \textbf{29.14}  & \textbf{1.14}    \\ \bottomrule
\end{tabular}
}
\caption{Transferring adversarial examples generated on source segmentation models to target models on PASCAL VOC 2012. We present the mIoU of target models on adversarial examples and corresponding clean images. For each source model, three adversarial attack methods, which include PGD, SegPGD, and our proposed attack, are used to generate adversarial examples. The mIoU of target models on the adversarial examples is lower, which indicates that the generated adversarial examples are easier to transfer. }
\label{table:voc_ours}
\end{table*}
\subsection{The second-stage adversarial attack strategy}
During the second stage, the goal is to boost the transferability of adversarial examples generated in the first stage, which could not only fool the source model $f^{seg}_{\boldsymbol{\theta}_{s}}$, but also the target model $f^{seg}_{\boldsymbol{\theta}_{t}}$. Motivated by the previous works of out-of-distribution, the well-trained models make it hard to identify the out-of-distribution examples. Several works have proven that the adversarial examples, which are further distributedly from the original clean examples, could have higher adversarial transferability. Image segmentation is an extension of image classification. Each pixel of the input image in the segmentation task can be considered as a sample in the classification task. Hence, generalized to the task of semantic segmentation, the adversarial pixels, which are farther distributedly from the original clean pixels, could also have higher adversarial transferability. Thus, we compute the Kullback-Leribler divergence between the generated adversarial pixels $\mathbf{x}_{adv}$ and the corresponding benign pixels $\mathbf{x}$. It can be calculated as:
\begin{equation}
D_{KL}(\mathbf{x}_{adv})^i = \sum_{j=1}^n \sigma( f^{seg}_{\boldsymbol{\theta}}(\mathbf{x}_{adv})^i)_j \log \frac{\sigma( f^{seg}_{\boldsymbol{\theta}}(\mathbf{x}_{adv})^i)_j}{\sigma( f^{seg}_{\boldsymbol{\theta}}(\mathbf{x})^i)_j},
\end{equation}
where $D_{KL}(\mathbf{x}_{adv})^i$ represents the KL distance between the model output on the $i$-th adversarial pixel and the model output on the corresponding clean pixel, $n$ represents the output dimension of the segmentation model, and $\sigma$ represents the softmax operation. Then we calculate the mean KL distance between all adversarial pixels and clean pixels, which can be formulated as $D_{KL}(\mathbf{x}_{adv})^{mean}$. We adopt the mean KL distance to divide the adversarial pixel into different branches,  \emph{ i.e.,} the high-transferability adversarial pixels $P_{H}$ and low-transferability adversarial pixels $P_{L}$. In detail, if the KL distance of the $i$-th pixel $D_{KL}(\mathbf{x}_{adv})^i$ is greater than the KL mean distance $D_{KL}(\mathbf{x}_{adv})^{mean}$, then the $i$-th adversarial pixel belongs to the high-transferability adversarial pixels $P_{H}$, otherwise it belongs to the low-transferability adversarial pixels $P_{L}$.
\par Hence, the loss function at the second stage can be formulated:
\begin{equation}
\begin{aligned}
\mathcal{L}\left(f^{seg}_{\boldsymbol{\theta}} \left(\mathbf{x}_{adv}^{t}\right), \mathbf{y}\right) &= \frac{1-\beta}{H \times W} \sum_{i \in P_H} \mathcal{L}_i\left(f^{seg}_{\boldsymbol{\theta}} \left(\mathbf{x}_{adv}^{t}\right), \mathbf{y}\right) \\ &+\frac{\beta}{H \times W} \sum_{j \in P_L} \mathcal{L}_j\left(f^{seg}_{\boldsymbol{\theta}} \left(\mathbf{x}_{adv}^{t}\right), \mathbf{y}\right),
\end{aligned}
\end{equation}
where $\beta$ represents a hyper-parameter at the second stage, which controls the allocation of weights. Based on the first-stage and second-stage adversarial attack strategies, we can establish our two-stage adversarial attack method to generate adversarial examples for semantic segmentation.  The algorithm
of the proposed method is summarized in Algorithm \ref{alg:attack}.
\section{Experiments}

\begin{table*}[t]
\centering

 \scalebox{0.85}{
 \centering
\begin{tabular}{@{}c|c|c|c|c|c|c|c@{}}
\toprule
Method       & PSPNet-Res50  & PSPNet-Res101  & DeepLabV3-Res50 & DeepLabV3-Res101 & FCNs-VGG       & Segformer & Segmenter \\ \midrule
Clean      & 78.55          & 79.11         & 78.17           & 80.55            & 69.10          &   77.19        &  78.5        \\ \midrule \midrule
MI-FGSM      & 5.46          & 29.71          & 5.85            & 32.63            & 36.88          &   43.61        &  54.53         \\ \midrule
MI-FGSM-ours & \textbf{2.09} & \textbf{26.42} & \textbf{2.62}   & \textbf{28.61}   & \textbf{34.62} &    \textbf{42.60}       &     \textbf{53.76}      \\ \midrule
TI-FGSM      & 4.89          & 35.43          & 7.08            & 39.60            & 35.62          &   47.30        &   55.42        \\ \midrule
TI-FGSM-ours & \textbf{2.19} & \textbf{35.06} & \textbf{5.12}   & \textbf{38.76}   & \textbf{34.88} &     \textbf{47.15}      &  \textbf{54.72}        \\ \midrule
NI-FGSM      & 5.59          & 30.18          & 5.96            & 32.67            & 37.02          &  43.82         &   54.50        \\ \midrule
NI-FGSM-ours & \textbf{2.17} & \textbf{26.35} & \textbf{2.77}   & \textbf{29.02}   & \textbf{34.69} &   \textbf{42.85}        &     \textbf{53.96}      \\ \bottomrule
\end{tabular}
}
\caption{Transferring adversarial examples generated on source segmentation models to target models on PASCAL VOC 2012. We present the mIoU of target models on adversarial examples and corresponding clean images. PSPNet-Res50 is used as the source model.
The mIoU of target models on the adversarial examples is lower, which indicates that the generated adversarial examples are easier to transfer. }
\label{table:voc_transfer}
\end{table*}
\begin{figure*}[t]
\begin{center}
 \includegraphics[width=1.0\linewidth]{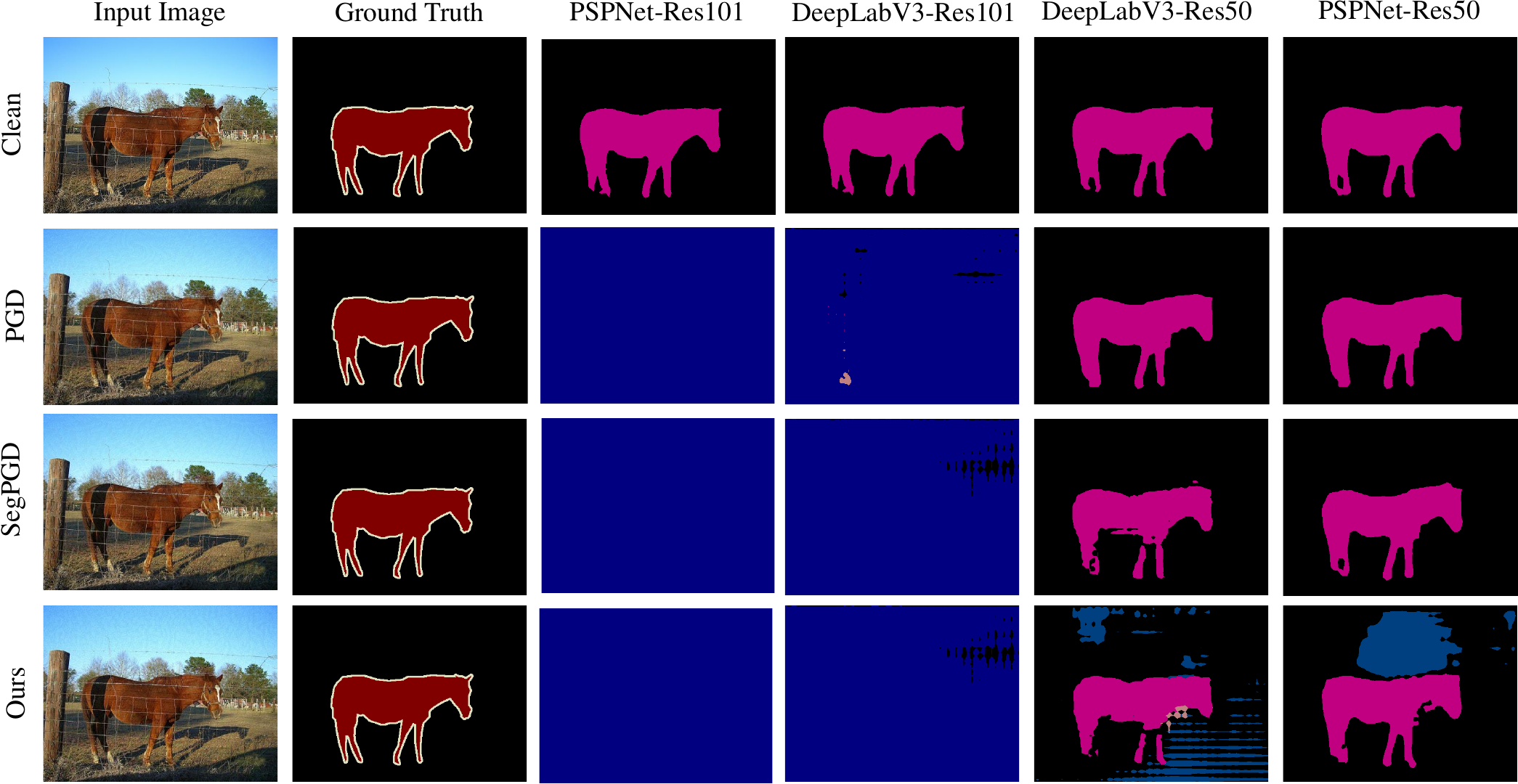}
\end{center}
\vspace{-3mm}
  \caption{ Visualization of Clean Images, Adversarial examples, and Segmentation Predictions. PSPNet-Res101 is used as the
source model. DeepLabV3-Res101, DeepLabV3-Res50, and PSPNet-Res50 are used as the
target models. The adversarial examples are generated by PGD, SegPGD, and the proposed TranSegPGD. 
The adversarial example generated by using the proposed method transfers better to other segmentation models. More figures are presented in the supplementary material.   }
\label{fig:vis}
\end{figure*}
\subsection{Settings}
Following previous works, we adopt widely used semantic segmentation datasets which include PASCAL VOC 2012
(VOC)~\cite{everingham2010pascal} and Cityscapes (CS)~\cite{cordts2016cityscapes} to conduct experiments. The VOC dataset consists of 20 classes for 
 objects and 1 class for background. It has 1,464 training images, 1,499 validation images, and 1,456 testing images. The Cityscapes dataset consists of Urban street scene images containing high-quality pixel-level annotations. It has 19 classes with  2,975 training images, 500 validation images, and 1,525 testing images. As for the semantic segmentation models, we use FCN8s-VGG16~\cite{shelhamer2017fully}, FCN16s-VGG16~\cite{shelhamer2017fully}, PSPNet-Res50~\cite{zhao2017pyramid}, PSPNet-Res101~\cite{zhao2017pyramid}, DeepLabv3-Res50~\cite{chen2017rethinking} and DeepLabv3-Res101~\cite{chen2017rethinking} to conduct adversarial example generation and performance evaluation. As for the baseline adversarial attack methods, we adopt the popular PGD and the advanced SegPGD. We also compare the proposed method with some popular transferable adversarial attack methods on image classification tasks which include TI-FGSM, MI-FGSM, and NI-FGSM to evaluate the effectiveness of the proposed method. All comparison experiments are under the $l_{\infty}$-norm. Specifically, we set the maximum perturbation strength $\epsilon$ to $8/255$, the attack step size $\alpha$ to $2/255$, and the number of attack iterations to 20. The mean Intersection over Union (mIoU) is used as a metric to evaluate the adversarial performance. 
 \begin{table*}[t]
\centering
 \scalebox{0.98}{
\begin{tabular}{c|c|c|c|c|c|c}
\toprule
Method       & PSPNet-Res50  & PSPNet-Res101  & DeepLabV3-Res50 & DeepLabV3-Res101 & Bisenet        & Segformer      \\ \midrule
Clean     & 74.20          &  76.04          & 74.06            & 76.05            & 75.16          & 81.08          \\ \midrule \midrule
MI-FGSM      & 1.06          & 16.64          & 1.56            & 21.99            & 41.39          & 45.86          \\ \midrule
MI-FGSM-ours & \textbf{0.32} & \textbf{13.20}  & \textbf{1.36}   & \textbf{18.56}   & \textbf{37.85} & \textbf{43.95} \\ \midrule
TI-FGSM      & 1.08          & 14.25          & 1.84            & 21.53            & 34.72          & 49.20          \\ \midrule
TI-FGSM-ours & \textbf{0.11} & \textbf{12.19} & \textbf{1.67}   & \textbf{19.49}   & \textbf{32.66} & \textbf{47.72} \\ \midrule
NI-FGSM      & 1.07          & 16.70          & 1.38            & 21.87            & 40.16          & 45.85          \\ \midrule
NI-FGSM-ours & \textbf{0.34} & \textbf{13.01} & \textbf{1.33}   & \textbf{18.39}   & \textbf{37.9}  & \textbf{43.97} \\ \bottomrule
\end{tabular}
}
\caption{Transferring adversarial examples generated on source segmentation models to target models on Cityscapes. We present the mIoU of target models on adversarial examples and corresponding clean images. PSPNet-Res50 is used as the source model.
The mIoU of target models on the adversarial examples is lower, which indicates that the generated adversarial examples are easier to transfer. }
\centering
\label{table:cite_ours}
\end{table*}
 \subsection{Comparisons with other adversarial attack methods on semantic segmentation}
 We compare the proposed method with the previous popular PGD and advanced SegPGD to evaluate adversarial transferability. In detail, we adopt PSPNet-Res50, PSPNet-Res101, DeepLabV3-Res50, and DeepLabV3-Res101 as the source models to generate adversarial examples on VOC. The results are shown in Table~\ref{table:voc_ours}. Analyses are as follows. First, the proposed method outperforms other adversarial attack methods under all attack scenarios. 
 \par In particular, compared with the popular PGD, the proposed method not only improves the adversarial performance of adversarial examples generated on the source model but also boosts the transferable adversarial performance of adversarial examples on the target model. For example, when using the PSPNet-Res50 as the source model, the proposed method improves the adversarial accuracy of popular PGD by about 3.05\% and improves a transferability performance of PGD on the PSPNet-Res101 by about 2.34\%. Besides, the proposed method also achieves the best adversarial transferability on other target models. We attribute the improvements to assigning different weights to different branches of the input image. Second, compared with the advanced SegPGD, the proposed method also achieves better adversarial transferability under all attack scenarios though there is always a trade-off between adversarial performance and transferability performance. Compared with PGD, SegPGD could achieve the limit improvement of adversarial example transferability. But, the proposed method can significantly improve the transferability of adversarial examples. It is attributed to the proposed second-stage adversarial attack strategy, which assigns different weights to different transferable branches. For example, when using the DeepLabV3-Res50 as the source model, the proposed method boosts the transferability performance of SegPGD on the DeepLabV3-Res101 by about 2.28\%. 
 \par Moreover, we visualize the adversarial examples and the prediction results of the adversarial examples on the other semantic segmentation models. In detail, we adopt the PSPNet-Res101 as the source model to generate adversarial examples and the DeepLabV3-Res101, DeepLabV3-Res50, and PSPNet-Res50 as the target models to evaluate the adversarial transferability. The result is shown in Figure~\ref{fig:vis}. It is clear that the adversarial generated by using PGD, SegPGD, and the proposed method can successfully fool the source model. Adversarial examples generated by using PGD and SegPGD do not significantly affect the output of the target models, but the adversarial examples generated by using the proposed method can fool the target models, which demonstrates the effectiveness of the proposed method in improving adversarial example transferability for semantic segmentation.

  \subsection{Comparisons with other transferable attack methods on image classification}
To further evaluate the effectiveness of TranSegPGD, we first generalize popular transferable attack methods which include MI-FGSM, TI-FGSM, and NI-FGSM on the image classification to the semantic segmentation and compare the proposed method with them. Our method can be combined with these transferable attack methods as a plug-and-play component to improve the transferability of adversarial examples on semantic segmentation, \emph{i.e.,} MI-FSGM-ours, TI-FGSM-ours, and NI-FGSM-ours.
\par For PASCAL VOC 2012, we use the PSPNet-Res50 as the source model to generate adversarial examples and use the PSPNet-Res101, DeepLabV3-Res50, DeppLabV3-Res101, FCNs-VGG, Segformer, and Segmenter as the target models to evaluate the adversarial transferability. 
The results are shown in Table~\ref{table:voc_transfer}. Performance analyses are summarized as follows. First, the proposed three-attack method achieves better adversarial transferability than their base adversarial attack methods under all attack scenarios. For example, when using the DeppLabV3-Res101 as the target model, the previous MI-FGSM achieves a transferability performance of about 32.63\%, but the proposed MI-FGSM-ours achieves a transferability performance of about 28.61\%, which boosts the 
transferability performance of about 4.02\%.
The previous TI-FGSM achieves a transferability performance of about 39.60\%, but the proposed TI-FGSM-ours achieves a transferability performance of about 38.76\%, which boosts the 
transferability performance of about 0.84\%.
The previous NI-FGSM achieves a transferability performance of about 32.67\%, but the proposed MI-FGSM-ours achieves a transferability performance of about 29.02\%, which boosts the 
transferability performance of about 3.65\%.
It indicates that TranSegPGD can significantly improve the transferability of adversarial examples. 

\par For Cityscapes, we also adopt the PSPNet-Res50 as the source model to generate adversarial examples and adopt the  PSPNet-Res101, DeepLabV3-Res50, DeepLabV3-Res101, Bisenet, and Segformer as the target models to evaluate the adversarial transferability. The results are shown in Table~\ref{table:cite_ours}. We can observe a similar phenomenon on PASCAL VOC 2012. The proposed method can boost the adversarial transferability of the base adversarial attack methods under all attack scenarios.  
For example, when using the PSPNet-Res101 as the target model, 
the previous MI-FGSM obtains a transferability performance of about 16.64\%, while the proposed MI-FGSM-ours achieves a transferability performance of about 13.2\%, which boosts the 
transferability performance of about 3.44\%.
The previous TI-FGSM obtains a transferability performance of about 14.25\%, while the proposed TI-FGSM-ours achieves a transferability performance of about 12.19\%, which boosts the 
transferability performance of about 2.06\%.
The previous NI-FGSM obtains a transferability performance of about 16.70\%, while the proposed NI-FGSM-ours achieves a transferability performance of about 13.01\%, which boosts the 
transferability performance of about 3.69\%. The experimental results indicate that the proposed method can further boost the transferability of adversarial examples.


\subsection{Transfer to attack segment anything model}
\begin{figure}[t]
\begin{center}
 \includegraphics[width=1.0\linewidth]{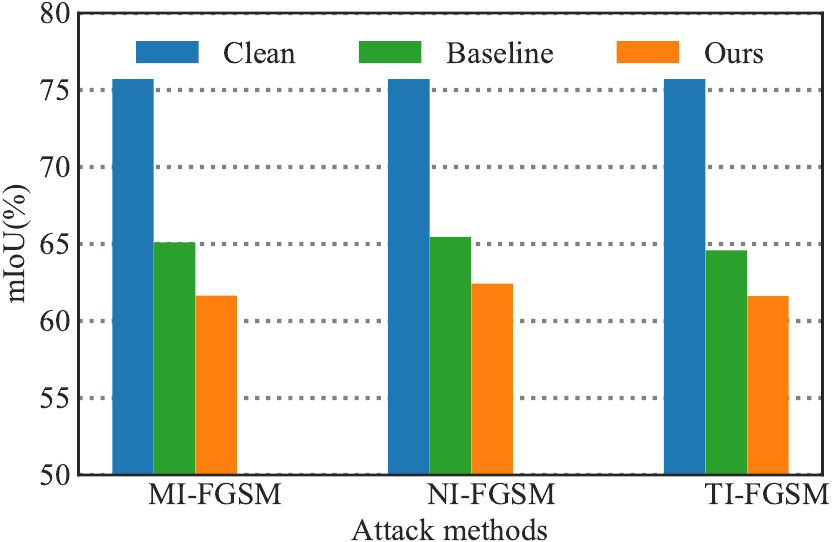}
\end{center}
\vspace{-6mm}
  \caption{ The mIoU of the segment anything model on adversarial examples generated on PSPNet-Res50 and corresponding clean image on the VOC. MI-FGSM, NI-FGSM, and TI-FGSM are used as the baseline models. $x$-axis represents
the attack methods. $y$-axis represents the mIoU(\%).  }
\label{fig:sam}
\vspace{-4mm}
\end{figure}

More and more works adopt foundation models to perform semantic segmentation tasks, and Segment Anything Model (SAM)~\cite{kirillov2023segment} stands out from them. 
We also transfer the adversarial examples to attack SAM to evaluate the effectiveness of the proposed method on the VOC. In detail, we adopt PSPNet-Res50 as the source model to generate adversarial examples. And MI-FGSM, TI-FGSM, and NI-FGSM are used as the baseline methods. The results are shown in Figure~\ref{fig:sam}. It can be observed that the proposed method can significantly improve the adversarial transferability performance of adversarial examples to SAM. We provide visualization results in the supplementary material. We also provide the experimental results on Cityscapes in the supplementary material.


\subsection{Ablation Study}
In this paper, we propose an effective two-stage adversarial attack strategy to boost the transferability of adversarial examples for semantic segmentation. The first-stage adversarial attack strategy is used to generate adversarial examples effectively. The second-stage adversarial attack strategy is used to improve the adversarial example transferability. To validate the
effectiveness of each stage in the proposed method, we
conduct ablation experiments on VOC. Specifically, we adopt the PSPNet-Res50 as the source model for adversarial example generation. PSPNet-Res101 and DeepLabV3-Res101 are used to validate the transferability of generated adversarial examples. The
results are shown in Table \ref{table:ablation}. Analyses are summarized as follows. 
\par First, when incorporating the first-stage adversarial attack strategy only, the adversarial performance on the source model significantly improves while the performance on the target models improves a little. When incorporating the second-stage adversarial attack strategy only, the adversarial performance on the source model slightly improves while the performance on the target models improves a lot. It indicates that the first-stage adversarial attack strategy contributes more to improve the adversarial performance and the second-stage adversarial attack strategy contributes more to improve the adversarial transferability. Second, using both adversarial attack strategies can achieve the adversarial performance on the source model and adversarial transferability performance, which suggests that the two adversarial attack strategies are harmonious, and their integration has the potential to achieve the best performance. 
\begin{table}[t]
\centering
 \scalebox{1}{
\begin{tabular}{@{}ccccc@{}}
\toprule
\begin{tabular}[c]{@{}c@{}}Fisrt\\ Stage\end{tabular} & \begin{tabular}[c]{@{}c@{}}Second\\ Stage\end{tabular} & \begin{tabular}[c]{@{}c@{}}PSPNet-\\ Res50\end{tabular} & \begin{tabular}[c]{@{}c@{}}PSPNet-\\ Res101\end{tabular} & \begin{tabular}[c]{@{}c@{}}DeepLabV3-\\ Res101\end{tabular} \\ \midrule
                                                    &                                                       & 4.60                                                    & 36.91                                                    & 38.59                                                       \\
\ding{52}                                                     &                                                       & 2.80                                                     & 36.38                                                    & 38.03                                                       \\
                                                    & \ding{52}                                                      & 3.93                                                    & 36.15                                                    & 37.22                                                       \\
\ding{52}                                                   & \ding{52}                                                     & \textbf{1.55}                                           & \textbf{34.57}                                           & \textbf{36.74}                                              \\ \bottomrule
\end{tabular}
}
\caption{Ablation study of the proposed method. The mIoU(\%) of segmentation models on the adversarial examples is reported. PSPNet-Res50 is used as the source model. PSPNet-Res101 and DeepLabV3-Res101 are used as the target models. }
\label{table:ablation}
\vspace{-6mm}
\end{table}

\section{Conclusion}
In this paper, we focus on how to improve the transferability of adversarial examples on semantic segmentation, which has been largely overlooked by previous works. We propose an effective two-stage adversarial attack strategy to boost transferability, \emph{dubbed } TranSegPGD. At the first stage, we divide each pixel in an input image into different branches according to its adversarial property. We assign distinct weights to different branches for optimization to enhance the adversarial performance of all pixels. At the second stage, we divide each pixel into different branches according to its transferable property, determined by Kullback-Leibler divergence. We assign distinct weights to different branches to boost transferability. We emphasize high weights on the loss of pixels with high transferability to amplify the transferability. Extensive experiments across diverse segmentation models conducted on the PASCAL VOC 2012 and Cityscapes datasets validate the efficacy of our method. The experiment results show that our TranSegPGD achieves state-of-the-art performance.
{
    \small
    \bibliographystyle{ieeenat_fullname}
    \bibliography{main}

\begin{thebibliography}{56}
\providecommand{\natexlab}[1]{#1}
\providecommand{\url}[1]{\texttt{#1}}
\expandafter\ifx\csname urlstyle\endcsname\relax
  \providecommand{\doi}[1]{doi: #1}\else
  \providecommand{\doi}{doi: \begingroup \urlstyle{rm}\Url}\fi

\bibitem[Arnab et~al.(2018)Arnab, Miksik, and Torr]{arnab2018robustness}
Anurag Arnab, Ondrej Miksik, and Philip~HS Torr.
\newblock On the robustness of semantic segmentation models to adversarial
  attacks.
\newblock In \emph{Proceedings of the IEEE conference on computer vision and
  pattern recognition}, pages 888--897, 2018.

\bibitem[Bai et~al.(2019)Bai, Feng, Wang, Dai, Xia, and Jiang]{bai2019hilbert}
Yang Bai, Yan Feng, Yisen Wang, Tao Dai, Shu-Tao Xia, and Yong Jiang.
\newblock Hilbert-based generative defense for adversarial examples.
\newblock In \emph{Proceedings of the IEEE/CVF International Conference on
  Computer Vision}, pages 4784--4793, 2019.

\bibitem[Bai et~al.(2020)Bai, Zeng, Jiang, Wang, Xia, and
  Guo]{bai2020improving}
Yang Bai, Yuyuan Zeng, Yong Jiang, Yisen Wang, Shu-Tao Xia, and Weiwei Guo.
\newblock Improving query efficiency of black-box adversarial attack.
\newblock In \emph{Computer Vision--ECCV 2020: 16th European Conference,
  Glasgow, UK, August 23--28, 2020, Proceedings, Part XXV 16}, pages 101--116.
  Springer, 2020.

\bibitem[Bai et~al.(2021{\natexlab{a}})Bai, Yan, Jiang, Xia, and
  Wang]{bai2021clustering}
Yang Bai, Xin Yan, Yong Jiang, Shu-Tao Xia, and Yisen Wang.
\newblock Clustering effect of (linearized) adversarial robust models.
\newblock \emph{arXiv preprint arXiv:2111.12922}, 2021{\natexlab{a}}.

\bibitem[Bai et~al.(2021{\natexlab{b}})Bai, Zeng, Jiang, Xia, Ma, and
  Wang]{bai2021improving}
Yang Bai, Yuyuan Zeng, Yong Jiang, Shu-Tao Xia, Xingjun Ma, and Yisen Wang.
\newblock Improving adversarial robustness via channel-wise activation
  suppressing.
\newblock \emph{arXiv preprint arXiv:2103.08307}, 2021{\natexlab{b}}.

\bibitem[Bai et~al.(2023)Bai, Wang, Zeng, Jiang, and Xia]{bai2023query}
Yang Bai, Yisen Wang, Yuyuan Zeng, Yong Jiang, and Shu-Tao Xia.
\newblock Query efficient black-box adversarial attack on deep neural networks.
\newblock \emph{Pattern Recognition}, 133:\penalty0 109037, 2023.

\bibitem[Chen et~al.(2017)Chen, Papandreou, Schroff, and
  Adam]{chen2017rethinking}
Liang-Chieh Chen, George Papandreou, Florian Schroff, and Hartwig Adam.
\newblock Rethinking atrous convolution for semantic image segmentation.
\newblock \emph{arXiv preprint arXiv:1706.05587}, 2017.

\bibitem[Cordts et~al.(2016)Cordts, Omran, Ramos, Rehfeld, Enzweiler, Benenson,
  Franke, Roth, and Schiele]{cordts2016cityscapes}
Marius Cordts, Mohamed Omran, Sebastian Ramos, Timo Rehfeld, Markus Enzweiler,
  Rodrigo Benenson, Uwe Franke, Stefan Roth, and Bernt Schiele.
\newblock The cityscapes dataset for semantic urban scene understanding.
\newblock In \emph{Proceedings of the IEEE conference on computer vision and
  pattern recognition}, pages 3213--3223, 2016.

\bibitem[Dong et~al.(2018)Dong, Liao, Pang, Su, Zhu, Hu, and
  Li]{dong2018boosting}
Yinpeng Dong, Fangzhou Liao, Tianyu Pang, Hang Su, Jun Zhu, Xiaolin Hu, and
  Jianguo Li.
\newblock Boosting adversarial attacks with momentum.
\newblock In \emph{Proceedings of the IEEE conference on computer vision and
  pattern recognition}, pages 9185--9193, 2018.

\bibitem[Dong et~al.(2019)Dong, Pang, Su, and Zhu]{dong2019evading}
Yinpeng Dong, Tianyu Pang, Hang Su, and Jun Zhu.
\newblock Evading defenses to transferable adversarial examples by
  translation-invariant attacks.
\newblock In \emph{Proceedings of the IEEE/CVF Conference on Computer Vision
  and Pattern Recognition}, pages 4312--4321, 2019.

\bibitem[Everingham et~al.(2010)Everingham, Van~Gool, Williams, Winn, and
  Zisserman]{everingham2010pascal}
Mark Everingham, Luc Van~Gool, Christopher~KI Williams, John Winn, and Andrew
  Zisserman.
\newblock The pascal visual object classes (voc) challenge.
\newblock \emph{International journal of computer vision}, 88:\penalty0
  303--338, 2010.

\bibitem[Fischer et~al.(2021)Fischer, Baader, and Vechev]{fischer2021scalable}
Marc Fischer, Maximilian Baader, and Martin Vechev.
\newblock Scalable certified segmentation via randomized smoothing.
\newblock In \emph{International Conference on Machine Learning}, pages
  3340--3351. PMLR, 2021.

\bibitem[Fischer et~al.(2017)Fischer, Kumar, Metzen, and
  Brox]{fischer2017adversarial}
Volker Fischer, Mummadi~Chaithanya Kumar, Jan~Hendrik Metzen, and Thomas Brox.
\newblock Adversarial examples for semantic image segmentation.
\newblock \emph{arXiv preprint arXiv:1703.01101}, 2017.

\bibitem[Frangi et~al.(2018)Frangi, Conjeti, Davatzikos, Navab, Schnabel,
  Alberola-L{\'o}pez, Fichtinger, Paschali, and
  Navarro]{frangi2018generalizability}
Alejandro~F Frangi, Sailesh Conjeti, Christos Davatzikos, Nassir Navab, Julia~A
  Schnabel, Carlos Alberola-L{\'o}pez, Gabor Fichtinger, Magdalini Paschali,
  and Fernando Navarro.
\newblock Generalizability vs. robustness: Investigating medical imaging
  networks using adversarial examples.
\newblock In \emph{International Conference on Medical Image Computing and
  Computer-Assisted Intervention}, number DZNE-2022-01068. Image Analysis,
  2018.

\bibitem[Goodfellow et~al.(2015)Goodfellow, Shlens, and
  Szegedy]{DBLP:journals/corr/GoodfellowSS14}
Ian~J. Goodfellow, Jonathon Shlens, and Christian Szegedy.
\newblock Explaining and harnessing adversarial examples.
\newblock In \emph{3rd International Conference on Learning Representations,
  {ICLR} 2015, San Diego, CA, USA, May 7-9, 2015, Conference Track
  Proceedings}, 2015.

\bibitem[Gu et~al.(2021)Gu, Zhao, Tresp, and Torr]{gu2021adversarial}
Jindong Gu, Hengshuang Zhao, Volker Tresp, and Philip Torr.
\newblock Adversarial examples on segmentation models can be easy to transfer.
\newblock \emph{arXiv preprint arXiv:2111.11368}, 2021.

\bibitem[Gu et~al.(2022)Gu, Zhao, Tresp, and Torr]{gu2022segpgd}
Jindong Gu, Hengshuang Zhao, Volker Tresp, and Philip~HS Torr.
\newblock Segpgd: An effective and efficient adversarial attack for evaluating
  and boosting segmentation robustness.
\newblock In \emph{European Conference on Computer Vision}, pages 308--325.
  Springer, 2022.

\bibitem[Gu et~al.(2023)Gu, Jia, de~Jorge, Yu, Liu, Ma, Xun, Hu, Khakzar, Li,
  et~al.]{gu2023survey}
Jindong Gu, Xiaojun Jia, Pau de Jorge, Wenqain Yu, Xinwei Liu, Avery Ma, Yuan
  Xun, Anjun Hu, Ashkan Khakzar, Zhijiang Li, et~al.
\newblock A survey on transferability of adversarial examples across deep
  neural networks.
\newblock \emph{arXiv preprint arXiv:2310.17626}, 2023.

\bibitem[He et~al.(2023{\natexlab{a}})He, Liu, Li, Liang, Li, Jia, and
  Cao]{he2023generating}
Bangyan He, Jian Liu, Yiming Li, Siyuan Liang, Jingzhi Li, Xiaojun Jia, and
  Xiaochun Cao.
\newblock Generating transferable 3d adversarial point cloud via random
  perturbation factorization.
\newblock In \emph{Proceedings of the AAAI Conference on Artificial
  Intelligence}, pages 764--772, 2023{\natexlab{a}}.

\bibitem[He et~al.(2023{\natexlab{b}})He, Zhang, Yang, He, Barnes, and
  Dai]{he2023transferable}
Mengqi He, Jing Zhang, Zhaoyuan Yang, Mingyi He, Nick Barnes, and Yuchao Dai.
\newblock Transferable attack for semantic segmentation.
\newblock \emph{arXiv preprint arXiv:2307.16572}, 2023{\natexlab{b}}.

\bibitem[He et~al.(2019)He, Yang, Li, Li, Chang, and Yu]{he2019non}
Xiang He, Sibei Yang, Guanbin Li, Haofeng Li, Huiyou Chang, and Yizhou Yu.
\newblock Non-local context encoder: Robust biomedical image segmentation
  against adversarial attacks.
\newblock In \emph{Proceedings of the AAAI Conference on Artificial
  Intelligence}, pages 8417--8424, 2019.

\bibitem[Hendrik~Metzen et~al.(2017)Hendrik~Metzen, Chaithanya~Kumar, Brox, and
  Fischer]{hendrik2017universal}
Jan Hendrik~Metzen, Mummadi Chaithanya~Kumar, Thomas Brox, and Volker Fischer.
\newblock Universal adversarial perturbations against semantic image
  segmentation.
\newblock In \emph{Proceedings of the IEEE international conference on computer
  vision}, pages 2755--2764, 2017.

\bibitem[Huang et~al.(2023{\natexlab{a}})Huang, Gao, and Liu]{huang2023erosion}
Lifeng Huang, Chengying Gao, and Ning Liu.
\newblock Erosion attack: Harnessing corruption to improve adversarial
  examples.
\newblock \emph{IEEE Transactions on Image Processing}, 2023{\natexlab{a}}.

\bibitem[Huang and Kong(2021)]{huang2021transferable}
Yi Huang and Adams Wai-Kin Kong.
\newblock Transferable adversarial attack based on integrated gradients.
\newblock In \emph{International Conference on Learning Representations}, 2021.

\bibitem[Huang et~al.(2023{\natexlab{b}})Huang, Cao, Li, Juefei-Xu, Lin, Tsang,
  Liu, and Guo]{huang2023robustness}
Yihao Huang, Yue Cao, Tianlin Li, Felix Juefei-Xu, Di Lin, Ivor~W Tsang, Yang
  Liu, and Qing Guo.
\newblock On the robustness of segment anything.
\newblock \emph{arXiv preprint arXiv:2305.16220}, 2023{\natexlab{b}}.

\bibitem[Jia et~al.(2020)Jia, Wei, Cao, and Han]{jia2020adv}
Xiaojun Jia, Xingxing Wei, Xiaochun Cao, and Xiaoguang Han.
\newblock Adv-watermark: A novel watermark perturbation for adversarial
  examples.
\newblock In \emph{Proceedings of the 28th ACM International Conference on
  Multimedia}, pages 1579--1587, 2020.

\bibitem[Jia et~al.(2022)Jia, Zhang, Wu, Ma, Wang, and Cao]{jia2022adversarial}
Xiaojun Jia, Yong Zhang, Baoyuan Wu, Ke Ma, Jue Wang, and Xiaochun Cao.
\newblock Las-at: adversarial training with learnable attack strategy.
\newblock In \emph{Proceedings of the IEEE/CVF Conference on Computer Vision
  and Pattern Recognition}, pages 13398--13408, 2022.

\bibitem[Kirillov et~al.(2023)Kirillov, Mintun, Ravi, Mao, Rolland, Gustafson,
  Xiao, Whitehead, Berg, Lo, et~al.]{kirillov2023segment}
Alexander Kirillov, Eric Mintun, Nikhila Ravi, Hanzi Mao, Chloe Rolland, Laura
  Gustafson, Tete Xiao, Spencer Whitehead, Alexander~C Berg, Wan-Yen Lo, et~al.
\newblock Segment anything.
\newblock \emph{arXiv preprint arXiv:2304.02643}, 2023.

\bibitem[LeCun et~al.(2015)LeCun, Bengio, and Hinton]{lecun2015deep}
Yann LeCun, Yoshua Bengio, and Geoffrey Hinton.
\newblock Deep learning.
\newblock \emph{nature}, 521\penalty0 (7553):\penalty0 436--444, 2015.

\bibitem[Li et~al.(2020)Li, Deng, Li, Yan, Gao, and Huang]{li2020towards}
Maosen Li, Cheng Deng, Tengjiao Li, Junchi Yan, Xinbo Gao, and Heng Huang.
\newblock Towards transferable targeted attack.
\newblock In \emph{Proceedings of the IEEE/CVF Conference on Computer Vision
  and Pattern Recognition}, pages 641--649, 2020.

\bibitem[Lin et~al.(2019)Lin, Song, He, Wang, and Hopcroft]{lin2019nesterov}
Jiadong Lin, Chuanbiao Song, Kun He, Liwei Wang, and John~E Hopcroft.
\newblock Nesterov accelerated gradient and scale invariance for adversarial
  attacks.
\newblock In \emph{International Conference on Learning Representations}, 2019.

\bibitem[Madry et~al.(2018)Madry, Makelov, Schmidt, Tsipras, and
  Vladu]{madry2018towards}
Aleksander Madry, Aleksandar Makelov, Ludwig Schmidt, Dimitris Tsipras, and
  Adrian Vladu.
\newblock Towards deep learning models resistant to adversarial attacks.
\newblock In \emph{International Conference on Learning Representations}, 2018.

\bibitem[Moosavi-Dezfooli et~al.(2016)Moosavi-Dezfooli, Fawzi, and
  Frossard]{moosavi2016deepfool}
Seyed-Mohsen Moosavi-Dezfooli, Alhussein Fawzi, and Pascal Frossard.
\newblock Deepfool: a simple and accurate method to fool deep neural networks.
\newblock In \emph{Proceedings of the IEEE conference on computer vision and
  pattern recognition}, pages 2574--2582, 2016.

\bibitem[Nagarajan et~al.(2020)Nagarajan, Andreassen, and
  Neyshabur]{nagarajan2020understanding}
Vaishnavh Nagarajan, Anders Andreassen, and Behnam Neyshabur.
\newblock Understanding the failure modes of out-of-distribution
  generalization.
\newblock In \emph{International Conference on Learning Representations}, 2020.

\bibitem[Qin et~al.(2022)Qin, Fan, Liu, Shen, Zhang, Wang, and
  Wu]{qin2022boosting}
Zeyu Qin, Yanbo Fan, Yi Liu, Li Shen, Yong Zhang, Jue Wang, and Baoyuan Wu.
\newblock Boosting the transferability of adversarial attacks with reverse
  adversarial perturbation.
\newblock \emph{Advances in Neural Information Processing Systems},
  35:\penalty0 29845--29858, 2022.

\bibitem[Shelhamer et~al.(2017)Shelhamer, Long, and
  Darrell]{shelhamer2017fully}
Evan Shelhamer, Jonathan Long, and Trevor Darrell.
\newblock Fully convolutional networks for semantic segmentation.
\newblock \emph{IEEE transactions on pattern analysis and machine
  intelligence}, 39\penalty0 (4):\penalty0 640--651, 2017.

\bibitem[Wald et~al.(2021)Wald, Feder, Greenfeld, and
  Shalit]{wald2021calibration}
Yoav Wald, Amir Feder, Daniel Greenfeld, and Uri Shalit.
\newblock On calibration and out-of-domain generalization.
\newblock \emph{Advances in neural information processing systems},
  34:\penalty0 2215--2227, 2021.

\bibitem[Wang and He(2021)]{wang2021enhancing}
Xiaosen Wang and Kun He.
\newblock Enhancing the transferability of adversarial attacks through variance
  tuning.
\newblock In \emph{Proceedings of the IEEE/CVF Conference on Computer Vision
  and Pattern Recognition}, pages 1924--1933, 2021.

\bibitem[Wang et~al.(2021)Wang, He, Wang, and He]{wang2021admix}
Xiaosen Wang, Xuanran He, Jingdong Wang, and Kun He.
\newblock Admix: Enhancing the transferability of adversarial attacks.
\newblock In \emph{Proceedings of the IEEE/CVF International Conference on
  Computer Vision}, pages 16158--16167, 2021.

\bibitem[Wang and Farnia(2023)]{wang2023role}
Yilin Wang and Farzan Farnia.
\newblock On the role of generalization in transferability of adversarial
  examples.
\newblock In \emph{Uncertainty in Artificial Intelligence}, pages 2259--2270.
  PMLR, 2023.

\bibitem[Wang et~al.(2023)Wang, Wang, Tian, and Jin]{wang2023adversarial}
Zhaoxin Wang, Handing Wang, Cong Tian, and Yaochu Jin.
\newblock Adversarial training of deep neural networks guided by texture and
  structural information.
\newblock In \emph{Proceedings of the 31st ACM International Conference on
  Multimedia}, pages 4958--4967, 2023.

\bibitem[Wu and Zhu(2020)]{wu2020towards}
Lei Wu and Zhanxing Zhu.
\newblock Towards understanding and improving the transferability of
  adversarial examples in deep neural networks.
\newblock In \emph{Asian Conference on Machine Learning}, pages 837--850. PMLR,
  2020.

\bibitem[Wu et~al.(2021)Wu, Su, Lyu, and King]{wu2021improving}
Weibin Wu, Yuxin Su, Michael~R Lyu, and Irwin King.
\newblock Improving the transferability of adversarial samples with adversarial
  transformations.
\newblock In \emph{Proceedings of the IEEE/CVF conference on computer vision
  and pattern recognition}, pages 9024--9033, 2021.

\bibitem[Xiao et~al.(2018)Xiao, Deng, Li, Yu, Liu, and
  Song]{xiao2018characterizing}
Chaowei Xiao, Ruizhi Deng, Bo Li, Fisher Yu, Mingyan Liu, and Dawn Song.
\newblock Characterizing adversarial examples based on spatial consistency
  information for semantic segmentation.
\newblock In \emph{Proceedings of the European Conference on Computer Vision
  (ECCV)}, pages 217--234, 2018.

\bibitem[Xie et~al.(2017)Xie, Wang, Zhang, Zhou, Xie, and
  Yuille]{xie2017adversarial}
Cihang Xie, Jianyu Wang, Zhishuai Zhang, Yuyin Zhou, Lingxi Xie, and Alan
  Yuille.
\newblock Adversarial examples for semantic segmentation and object detection.
\newblock In \emph{Proceedings of the IEEE international conference on computer
  vision}, pages 1369--1378, 2017.

\bibitem[Xie et~al.(2019)Xie, Zhang, Zhou, Bai, Wang, Ren, and
  Yuille]{xie2019improving}
Cihang Xie, Zhishuai Zhang, Yuyin Zhou, Song Bai, Jianyu Wang, Zhou Ren, and
  Alan~L Yuille.
\newblock Improving transferability of adversarial examples with input
  diversity.
\newblock In \emph{Proceedings of the IEEE/CVF conference on computer vision
  and pattern recognition}, pages 2730--2739, 2019.

\bibitem[Xiong et~al.(2022)Xiong, Lin, Zhang, Hopcroft, and
  He]{xiong2022stochastic}
Yifeng Xiong, Jiadong Lin, Min Zhang, John~E Hopcroft, and Kun He.
\newblock Stochastic variance reduced ensemble adversarial attack for boosting
  the adversarial transferability.
\newblock In \emph{Proceedings of the IEEE/CVF Conference on Computer Vision
  and Pattern Recognition}, pages 14983--14992, 2022.

\bibitem[Xu et~al.(2021)Xu, Zhao, and Jia]{xu2021dynamic}
Xiaogang Xu, Hengshuang Zhao, and Jiaya Jia.
\newblock Dynamic divide-and-conquer adversarial training for robust semantic
  segmentation.
\newblock In \emph{Proceedings of the IEEE/CVF International Conference on
  Computer Vision}, pages 7486--7495, 2021.

\bibitem[Yu et~al.(2023)Yu, Gu, Li, and Torr]{yu2023reliable}
Wenqian Yu, Jindong Gu, Zhijiang Li, and Philip Torr.
\newblock Reliable evaluation of adversarial transferability.
\newblock \emph{arXiv preprint arXiv:2306.08565}, 2023.

\bibitem[Zhang et~al.(2022)Zhang, Benz, Karjauv, Cho, Zhang, and
  Kweon]{zhang2022investigating}
Chaoning Zhang, Philipp Benz, Adil Karjauv, Jae~Won Cho, Kang Zhang, and In~So
  Kweon.
\newblock Investigating top-k white-box and transferable black-box attack.
\newblock In \emph{Proceedings of the IEEE/CVF Conference on Computer Vision
  and Pattern Recognition}, pages 15085--15094, 2022.

\bibitem[Zhang et~al.(2023)Zhang, Huang, Wang, Li, Wu, Wang, Su, and
  Lyu]{zhang2023improving}
Jianping Zhang, Jen-tse Huang, Wenxuan Wang, Yichen Li, Weibin Wu, Xiaosen
  Wang, Yuxin Su, and Michael~R Lyu.
\newblock Improving the transferability of adversarial samples by
  path-augmented method.
\newblock In \emph{Proceedings of the IEEE/CVF Conference on Computer Vision
  and Pattern Recognition}, pages 8173--8182, 2023.

\bibitem[Zhao et~al.(2017)Zhao, Shi, Qi, Wang, and Jia]{zhao2017pyramid}
Hengshuang Zhao, Jianping Shi, Xiaojuan Qi, Xiaogang Wang, and Jiaya Jia.
\newblock Pyramid scene parsing network.
\newblock In \emph{Proceedings of the IEEE conference on computer vision and
  pattern recognition}, pages 2881--2890, 2017.

\bibitem[Zhao et~al.(2021)Zhao, Liu, and Larson]{zhao2021success}
Zhengyu Zhao, Zhuoran Liu, and Martha Larson.
\newblock On success and simplicity: A second look at transferable targeted
  attacks.
\newblock \emph{Advances in Neural Information Processing Systems},
  34:\penalty0 6115--6128, 2021.

\bibitem[Zhou et~al.(2016)Zhou, Khosla, Lapedriza, Oliva, and
  Torralba]{zhou2016learning}
Bolei Zhou, Aditya Khosla, Agata Lapedriza, Aude Oliva, and Antonio Torralba.
\newblock Learning deep features for discriminative localization.
\newblock In \emph{Proceedings of the IEEE conference on computer vision and
  pattern recognition}, pages 2921--2929, 2016.

\bibitem[Zhu et~al.(2023)Zhu, Zheng, Zhu, and Sui]{zhu2023boosting}
Hegui Zhu, Haoran Zheng, Ying Zhu, and Xiaoyan Sui.
\newblock Boosting the transferability of adversarial attacks with adaptive
  points selecting in temporal neighborhood.
\newblock \emph{Information Sciences}, 641:\penalty0 119081, 2023.

\bibitem[Zhu et~al.(2022)Zhu, Chen, Li, Chen, He, Tian, Zheng, Chen, and
  Huang]{zhu2022toward}
Yao Zhu, Yuefeng Chen, Xiaodan Li, Kejiang Chen, Yuan He, Xiang Tian, Bolun
  Zheng, Yaowu Chen, and Qingming Huang.
\newblock Toward understanding and boosting adversarial transferability from a
  distribution perspective.
\newblock \emph{IEEE Transactions on Image Processing}, 31:\penalty0
  6487--6501, 2022.

\end{thebibliography}
}


\end{document}